\documentclass[10pt,twocolumn,letterpaper]{article}

\usepackage{cvpr}              % To produce the CAMERA-READY version
\usepackage{algorithm}
\usepackage{algorithmic}
\definecolor{cvprblue}{rgb}{0.21,0.49,0.74}
\usepackage[pagebackref,breaklinks,colorlinks,allcolors=cvprblue]{hyperref}
\usepackage{booktabs} 
\usepackage{multirow}
\usepackage{colortbl} 
\usepackage{xcolor}
\usepackage{courier}
\usepackage[frozencache,cachedir=.]{minted}
\usepackage{algorithm}
\usepackage{listings}

\definecolor{darkgreen}{rgb}{0.0, 0.5, 0.0}

\def\paperID{16724} % *** Enter the Paper ID here
\def\confName{CVPR}
\def\confYear{2025}

\title{Static Key Attention in Vision}

\author{
Zizhao Hu, Xiaolin Zhou, Mohammad Rostami \\
University of Southern California \\
{\tt\small \{zizhaoh, xzhou733, rostamim\}@usc.edu}
}

\begin{document}
\maketitle
\begin{abstract}
The success of vision transformers is widely attributed to the expressive power of their dynamically parameterized multi-head self-attention mechanism. We examine the impact of substituting the dynamic parameterized key with a static key within the standard attention mechanism in Vision Transformers. Our findings reveal that static key attention mechanisms can match or even exceed the performance of standard self-attention. Integrating static key attention modules into a Metaformer backbone, we find that it serves as a better intermediate stage in hierarchical hybrid architectures,   balancing the strengths of depth-wise convolution and self-attention. Experiments on several vision tasks underscore the effectiveness of the static key mechanism, indicating that the typical two-step dynamic parameterization in attention can be streamlined to a single step without impacting performance under certain circumstances. 
\end{abstract}    
\section{Introduction}
\label{sec:intro}

The Vision Transformer (ViT)~\cite{vit} and its variants have become integral to computer vision. ViT leverages the multi-head self-attention (MHSA)~\cite{MHSA} mechanism to capture complex dependencies between image patches. Since MHSA was initially developed for NLP, the dynamic dot products of query and key vectors are essential to handle global token dependencies with flexible context lengths. However, such dynamics may not be necessary in computer vision, where a static spatial image structure is exhibited across image datasets and biological vision. This property in vision raises the question of whether a single attention-value dynamic dot product could sufficiently capture spatial dependencies in image patches while keeping the query-key interaction static to reduce the complexity of the self-attention mechanism without performance degradation.

Recent works on vision frameworks such as MetaMixer \cite{metamixer} and MetaFormer \cite{metaformer} have explored the possibility of making MHSA more efficient. They have shown that the attention mechanism can be replaced with different blocks while maintaining model performance. These frameworks are reflected in various state-of-the-art architectures such as Xception~\cite{xception}, ConvMixer~\cite{convmixer}, and MLP-Mixer~\cite{mlpmixer}, which offer alternative methods for mixing tokens that are more computationally efficient than self-attention. We propose a remarkably different strategy to achieve this goal.
Instead of introducing vastly different new token mixing architecture, we focus on potential redundancies in the MHSA itself and simplify MHSA when the input is an image. 

We propose and investigate the static key mechanism in MHSA. Specifically, we introduce Static Key Attention (SKA) and Convolutional Static Key Attention (CSKA) by replacing the dynamic query-key attention with a static key mechanism following query calculation, while retaining the dynamic attention-value operation. SKA directly replaces the dynamic key in place, while CSKA explores the connection between SKA and grouped convolutions, offering a hybrid model that leverages the strengths of both architectures. SKA and CSKA aim to study how a retrieval-like query-key interaction can behave differently than the standard dynamic interaction and its potential benefits.

We empirically validate the effectiveness of SKA and CSKA through extensive experiments on various vision tasks, including image classification, object detection, and segmentation. Our results indicate that these architectures can match or, under specific conditions, outperform standard MHSA or depth-wise attention in terms of parameter efficiency and inference speed. By bridging the gap between convolutional and self-attention-based models, SKA and CSKA provide alternative approaches to traditional MHSA-based architectures with more resource efficiency in certain scenarios. Our contributions include:

\begin{itemize} 

\item We present insights on how static key affects the performance of attention mechanism in vision tasks by introducing two new static key attention block architectures, SKA and CSKA. 
\item We offer insights into how these architectures behave compared to the standard attention mechanism and present pros and cons and explore them empirically. 
\item Through comprehensive experiments on various vision tasks, we demonstrate that SKA and CSKA achieve competitive performance similar to the multi-head attention mechanism and, when used as an intermediate layer in a hierarchical architecture, achieve SOTA performance. \end{itemize}

\section{Efficient Architectures for Vision}
\label{sec:related}

Transformer was originally designed for machine translation and later adopted in  vision domain as ViT\cite{vit}. By breaking down images into patch tokens, the MHSA mechanism can be directly used to learn complex spatial interactions in image data. However, the high computational and memory costs of ViTs, especially for high-resolution images and deep models limit their practical adoption. To address these issues, subsequent research has proposed adaptations to make the standard ViT less computationally demanding, such as local attention networks\cite{swin, halo, cswin, nat}, deep hierarchical networks~\cite{cvt, pvt, metamixer, metaformer, nat}, and hybrid architectures~\cite{cvt, metamixer, coatnet}. 

Another line of work is mixer models which try to relax the need for MHSA using other mechanism. Although designed as attention-free models, many variants follow the design of Transformers. More recent studies generalize such models under the Metaformer~\cite{metaformer} and Metamixer~\cite{metamixer} frameworks, with Metaformer being the framework we mainly use in this paper for experimentation. The Metaformer framework demonstrates that we can create ViT-like architectures with alternative token-mixing modules other than the MHSA, such as depth-wise convolutions and MLPs. Thus we consider these mixer models as special ViTs that explored the possibility of reducing the complexity of MHSA by replacing the attention module with simpler token mixing networks~\cite{convmixer, mlpmixer, gmlp, resmlp}.

Another research direction has explored alternative structures within the MHSA to lower computational costs while preserving the attention mechanism. One notable approach is sparse attention, where attention computation is restricted to specific regions or subsets of tokens instead of all pairs. For instance, Vision Longformer~\cite{longformer} introduces a combination of local and global attention patterns to reduce complexity, making it suitable for longer sequences. Another example is Linformer~\cite{linformer}, which approximates the attention matrix by projecting keys and values into lower-dimensional spaces. Similarly, Performer~\cite{choromanski2020rethinking} uses kernel-based approximations to replace softmax attention. These methods demonstrate the possibility of modifying MHSA to improve efficiency while maintaining performance.

In contrast to the above, we focus on modifying the attention mechanism itself by making it more efficient. While most of the above works keep the basic query-key-value operation intact, we study the effect of replacing the dynamic key-query mechanism in MHSA with a trainable static key while retaining dynamic weights between the attention score and value interaction. Our approach reduces the overhead of dynamically generating keys, yet it preserves the dynamic parameterization of MHSA. In introducing SKA and CSKA, our work provides insight into a static key attention mechanism, offering an alternative architecture that under certain conditions such as a small dataset or used with other token mixers in a hierarchical network, proves efficient and scalable. Unlike prior works that often rely on predefined or pretrained static keys, our method is fully end-to-end because the static key is learned directly from data during training and   is more adoptable. 

\section{Background}

\paragraph{Multi-Head Self-Attention:} 
MHSA enables each token to attend to all others in an input in the form of a sequence, capturing global dependencies effectively. Given an input sequence $X \in \mathbb{R}^{N \times D}$ of $N$ tokens with embedding dimension $D$, MHSA computes three linear projections for Queries ($Q$), Keys ($K$), and Values ($V$) using learnable weights $W_Q, W_K, W_V \in \mathbb{R}^{D \times D_h}$, where $D_h = D / H$ and $H$ is the number of attention heads:
\[
Q = X W_Q, \quad K = X W_K, \quad V = X W_V.
\]
The attention scores are computed via scaled dot-product:
\[
\text{Attention}(Q, K, V) = \text{softmax}\left(\frac{QK^\top}{\sqrt{D_h}}\right)V,
\]
producing outputs for each head. Outputs from all heads are concatenated and linearly projected back to $D$.

\paragraph{MetaFormer:}
It is a general framework for understanding and designing token-mixing mechanisms, abstracting the process of token mixing into a flexible, modular architecture ~\cite{metaformer}.  Metaformer is not limited to self-attention as is used in ViT; instead, it allows for a variety of token-mixing operations, such as convolutions, multi-layer perceptrons (MLPs), or simple random mixing. 

\paragraph{Token Mixers:} They are an alternative class of models that aim to capture spatial dependencies without the full complexity of self-attention. Many token mixers are used in conjunction with channel mixers to create expressive models such as ConvMixer~\cite{convmixer} and MLP-Mixer~\cite{mlpmixer}, which use convolutional and fully connected layers for token mixing respectively. These architectures prioritize efficiency by replacing self-attention with simpler operations, making them computationally attractive for large-scale image processing tasks. Various token mixers have demonstrated competitive performance on certain vision tasks, with MetaFormer unifying these alternatives as potential replacements for self-attention layers.

\paragraph{Depth-wise and Grouped Convolutions:}
Convolutional neural networks (CNNs) remain highly competitive in terms of computational efficiency and inference speed in the vision domain. Their ability to exploit local patterns through hierarchical convolutional feature extraction makes them efficient in preprocessing high-resolution images. However, due to the weight sharing, CNNs struggle to achieve comparable expressiveness in complex vision tasks like ViTs. Hence, CNNs often remain the preferred choice for the early processing of high-resolution data, compressing, and extracting features. Introduced first in the Xception model~\cite{xception}, depth-wise convolutions reduce computational complexity by applying a single convolutional filter per input channel (depth-wise convolution) followed by a pointwise convolution to combine channel-wise information. Depth-wise convolution allows efficient spatial mixing with fewer parameters compared to standard convolutions. It has also been adapted in mixer models as efficient alternatives to standard convolutions or MHSA, particularly in early layers where local spatial mixing is more critical~\cite{convmixer,metaformer}. Grouped convolutions, on the other hand, divide the input channels into groups, with separate convolutions applied to each group. The standard convolution and depth-wise convolution can be seen as special cases of grouped convolution, when the number of groups is equal to 1 or the number of input channels, respectively. Grouped convolution allows more fine-grained control of complexity between standard convolution and depth-wise convolution. 

In this work, we investigate the impact of static keys within the MHSA framework. Specifically, we propose Static Key Attention (SKA) and Convolutional Static Key Attention (CSKA), which introduce static keys to MHSA in ways inspired by both transformer and convolutional paradigms. Our approach aims to reduce the computational burden of MHSA while retaining the flexibility of dynamic weight adaptation, drawing insights from both ViTs and convolutional token-mixing techniques.

\section{Proposed Method}

\begin{figure}
    \centering
    \includegraphics[width=\linewidth]{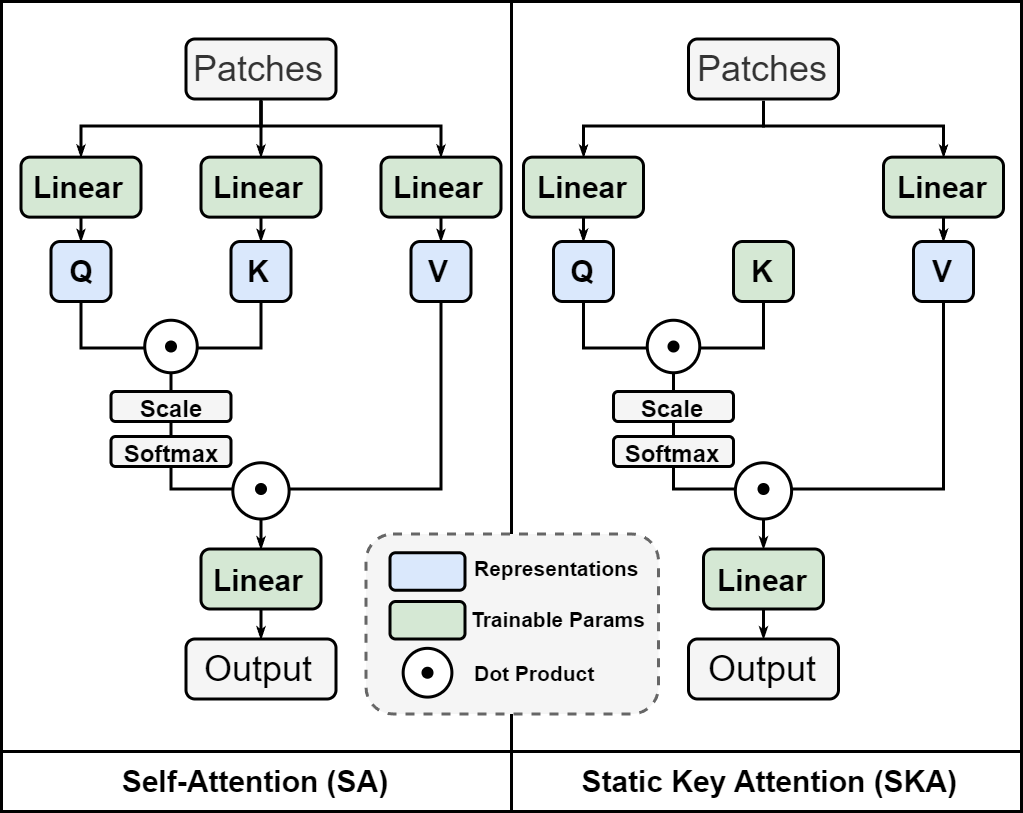}
    \caption{The standard self-attention mechanism in (left) Vision Transformers and (right) Static Key Attention, where it replaces the linear layer generating the key tensor with a fixed, learnable key matrix, removing the need for extra transformations.}
    \label{fig:SASKA}
\end{figure}

% We introduce two approaches to incorporate a static key into  MHSA. 

We use either multi-head fully connected layers or grouped convolutions to replace the original query-key interaction in MHSA,  forming the basis for Static Key Attention (SKA) and Convolutional Static Key Attention (CSKA). See the Appendix for code implementation.

\subsection{Static Key Attention (SKA)}

\begin{figure*}
    \centering
    \includegraphics[width=1\linewidth]{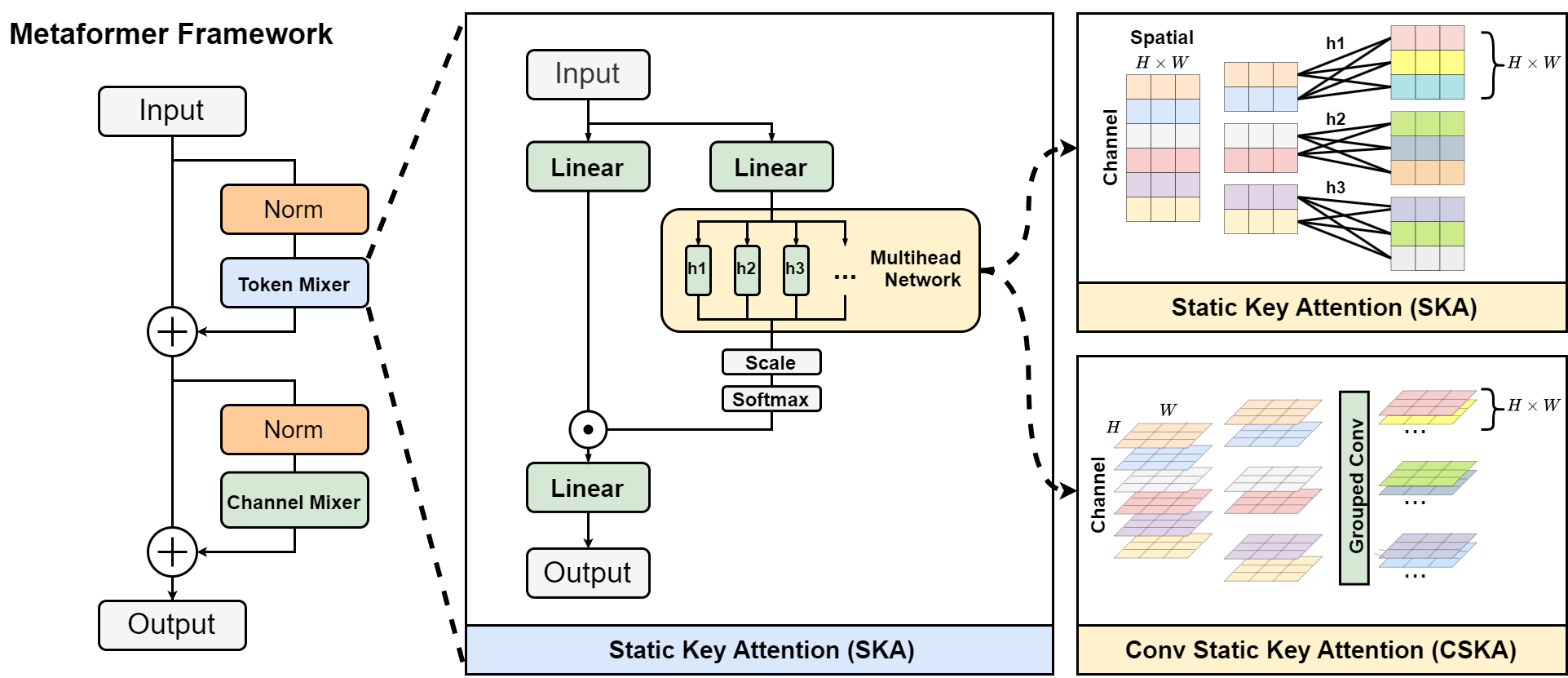}
    \caption{The Static Key Attention (SKA) module and Convolutional Static Key Attention module (CSKA) we proposed are used  under the MetaFormer framework. They serve as new types of token mixers, while the channel mixer is an MLP which is identical to other MetaFormer modules, such as the Vision Transformer. They can also be adopted in other architectures.}
    \label{fig:pipeline}
\end{figure*}

\textbf{Static Key Attention (SKA)} modifies the standard MHSA mechanism by introducing a static weight matrix \( W_k \) per attention head (Figure \ref{fig:SASKA} shows how we can modify self-attention to SKA, Figure \ref{fig:pipeline} top right shows a more detailed implementation under the MetaFormer framework). This static matrix transforms the channel dimension of the input tensor into the spatial dimension.

\paragraph{Static Key Mechanism:}

We replace the dynamic key \( K \) with a learned static weight matrix \( W_K \) for each head. The query-key interaction in MHSA with \( H \) heads becomes:
\[
A_h = Q_h \cdot W_{K,h}^T \quad \text{for each head} \ h = 1, 2, \dots, H,
\]
where \( Q_h \in \mathbb{R}^{N \times d_h} \) is the query matrix for head \( h \), \( W_{K,h} \in \mathbb{R}^{d_h \times N} \) is the static weight matrix for head \( h \), and \( N \) is the number of tokens. Figure \ref{fig:SASKA} shows a visualization of the computational pipeline of SKA vs standard self-attention. The multi-head static attention map \( A_h \) is then computed by concatenating the attention maps across all heads as:
\[
A = \text{Concat}(A_1, A_2, \dots, A_H).
\]

\paragraph{Dynamic Transformation:}
Similar to standard MHSA, upon calculating the static attention map \( A_h \), the output of the multi-head attention is computed by using the attention map to transform the value matrix \( V_h \) for each head:
\[
O_h = \text{Softmax}(A_h) \cdot V_h,
\]
where \( V_h \in \mathbb{R}^{B \times N \times d_h} \). The final output of SKA is obtained by concatenating the outputs from all heads and projecting them back to the original feature space as:
\[
\text{SKA}(X) = \text{Concat}(O_1, O_2, \dots, O_H) \cdot W_O,
\]
where \( W_O \in \mathbb{R}^{H \cdot d_h \times d} \) is the projection matrix that merges the outputs from the different heads into the final output.

\subsection{Convolutional Static Key Attention (CSKA)}

CSKA introduces grouped convolutions to generate the static weight matrix \( W_k \) (Figure \ref{fig:pipeline} bottom right shows a detailed implementation under the MetaFormer framework). In CSKA, the number of groups in the convolution corresponds to the number of heads, allowing each head to independently transform its channels into a spatial representation which produces a square \( N \times N \) attention matrix, where \( N \) is the number of input tokens.

\paragraph{Grouped Convolution Static Key:}

The grouped convolution directly produces the attention map \( A_h \) for each head \( h \) from the query projections, expressed as:
\[
A_h = \text{GroupedConv}(Q, h),
\]
where \( Q \in \mathbb{R}^{B \times N \times d} \) is the input query, calculated in the same way as in standard attention. Here, \( h \) is the number of groups (heads), and \( N \) represents the spatial dimensions (flattened height and width). Each group outputs spatial dimension channels, resulting in a total of \( h \times N \) output channels.To maintain the spatial shape, we use a \( 3 \times 3 \) kernel, stride \( 1 \), and padding \( 1 \). The result is a square attention map \( A_h \in \mathbb{R}^{N \times N} \) for each head, where \( N \) is the number of spatial tokens.
This design directly leverages the grouped convolution to compute \( A_h \), avoiding the need for additional dynamic query-key interactions.

\paragraph{Dynamic transformation:}

Same as SKA, the dynamic transformation is applied to the value for each group after calculating the attention map \( A_h \).

\subsection{Comparison with related block designs}

\paragraph{Multi-head self-attention:}
MHSA dynamically computes the key, query, and value matrices. SKA and CSKA do not compute the key dynamically. Instead, a network is used as the key. The key network is designed to ensure the resulting query-key attention map remains a square matrix.

\paragraph{Depth-wise convolution:}
Depth-wise convolutions can be seen as a special case of grouped convolution when the number of groups is set equal to the number of channels, and the output number of channels is equal to the input number of channels. We use the more general grouped convolution in CSKA which helps to balance the performance and speed. CSKA changes the multi-head query-key interaction of MHSA into grouped convolutions, where the groups correspond to the heads in MHSA.  

\paragraph{Gated Linear Unit (GLU):}
GLU introduces a gating mechanism to offer an attention mechanism by controlling information flow between different pathways, which enhances the model’s expressive power by selectively passing or blocking relevant features. SKA and CSKA share a similar two-pathway structure but they use a dot-product gate, and the activation is replaced by a scaled softmax function.

\section{Comparative experiments}
To demonstrate the effectiveness of SKA and CSKA, we conducted two sets of experiments on \textit{classification} and \textit{object detection with instance segmentation}. For classification, the experiments were performed on small-scale datasets (CIFAR-10 and CIFAR-100, additional datasets see the Appendix) and a larger-scale dataset (ImageNet). For object detection and instance segmentation, we followed the setup used in ~\cite{metaformer} and conducted experiments on the COCO dataset. For details on implementation see the Appendix. Our implementation is also available as a supplement.
 
\subsection{Comparative results for image classification}
\begin{table}[h]
\scriptsize
\centering
\caption{Comparative result on CIFAR: SKA/CSKA models are constructed by replacing the standard attention mechanism with SKA/CSKA in each of the models.}
\label{tab:model_comparison_cifar}
\begin{tabular}{lcccl}
\toprule
\textbf{Model} & \textbf{\#Params} & \textbf{FLOPs} & \textbf{Tp (image/s)} & \textbf{Top-1 Acc.} \\
\midrule
\multicolumn{5}{c}{\textbf{CIFAR-10}} \\
\midrule
CvT-13 & 19.6M & 0.17G & 6349 & 81.0 \\
\rowcolor{gray!10}
&18.0M& 0.21G & 6896 & 81.3 \textcolor{darkgreen}{(+0.3)} \\
\rowcolor{gray!10}
\multirow{-2}{*}{\textbf{SKA-CvT-13}}&19.6M& 0.22G & 6440 & 81.4 \textcolor{darkgreen}{(+0.4)} \\
\rowcolor{gray!10}
\textbf{CSKA-CvT-13}&18.0M& 0.19G & 7042 & \textbf{81.3} \textcolor{darkgreen}{(+0.3)} \\
\midrule
ViT-S & 9.6M & 1.29G & 5008 & 83.4 \\
\rowcolor{gray!10}
& 8.1M & 1.25G & 5120 & 84.1 \textcolor{darkgreen}{(+0.7)} \\
\rowcolor{gray!10}
\multirow{-2}{*}{\textbf{SKA-ViT-S }}& 9.5M & 1.28G & 5010 &84.7 \textcolor{darkgreen}{(+1.3)} \\
\rowcolor{gray!10}
\textbf{CSKA-ViT-S }& 10.0M & 1.33G & 7220 & \textbf{87.0} \textcolor{darkgreen}{(+3.6)} \\
\midrule
CaiT-S& 12.7M & 1.42G & 7358 & 82.8 \\
\rowcolor{gray!10}
&11.3M & 1.37G & 7880 & 83.5 \textcolor{darkgreen}{(+0.7)} \\
\rowcolor{gray!10}\multirow{-2}{*}{\textbf{SKA-CaiT-S }}& 12.7M & 1.41G & 6830 & 84.0 \textcolor{darkgreen}{(+1.2)} \\
\rowcolor{gray!10}\textbf{CSKA-CaiT-S }& 12.8M & 1.44G & 7575 & \textbf{87.9} \textcolor{darkgreen}{(+5.1)} \\
\midrule
Swin-T & 26.6M & 1.05G & 5105 & 89.2 \\
\rowcolor{gray!10}
& 24.6M & 0.97G & 5523 & 89.7 \textcolor{darkgreen}{(+0.5)} \\
\rowcolor{gray!10}
\multirow{-2}{*}{\textbf{SKA-Swin-T }}& 26.7M & 0.99G & 5200 & \textbf{90.0} \textcolor{darkgreen}{(+0.8)} \\
\rowcolor{gray!10}
\textbf{CSKA-Swin-T }& 26.8M & 1.52G & 4949 & 89.4 \textcolor{darkgreen}{(+0.2)} \\
\midrule
\multicolumn{5}{c}{\textbf{CIFAR-100}} \\
\midrule
CvT-13& 19.6M & 0.17G & 6410 & 49.8 \\
\rowcolor{gray!10}
&18.0M& 0.21G & 6859 & 49.8 \textcolor{darkgreen}{(0.0)} \\
\rowcolor{gray!10}
\multirow{-2}{*}{\textbf{SKA-CvT-13}}&19.6M& 0.22G & 6443 & 50.5 \textcolor{darkgreen}{(+0.7)} \\
\rowcolor{gray!10}
\textbf{CSKA-CvT-13}&18.0M& 0.19G & 7042 & \textbf{51.1} \textcolor{darkgreen}{(+0.9)} \\
\midrule
CaiT-S& 12.7M & 1.42G & 7434 & 57.3 \\
\rowcolor{gray!10}
&11.3M & 1.37G & 7757 & 57.4 \textcolor{darkgreen}{(+0.1)} \\
\rowcolor{gray!10}
\multirow{-2}{*}{\textbf{SKA-CaiT-S}}& 12.7M & 1.41G & 6915 & 57.8 \textcolor{darkgreen}{(+0.5)} \\
\rowcolor{gray!10}
\textbf{CSKA-CaiT-S}& 12.8M & 1.44G & 7407 & \textbf{60.6} \textcolor{darkgreen}{(+3.3)} \\
\midrule
Swin-T & 26.6M & 1.05G & 5105 & 63.1 \\
\rowcolor{gray!10}
& 24.6M & 0.97G & 5523 & 62.2 \textcolor{red}{(-0.9)} \\
\rowcolor{gray!10}
\multirow{-2}{*}{\textbf{SKA-Swin-T }}& 26.8M & 0.99G & 5200 & \textbf{63.3} \textcolor{darkgreen}{(+0.2)} \\
\rowcolor{gray!10}
\textbf{CSKA-Swin-T } & 26.8M & 1.52G & 4949 & 62.3 \textcolor{red}{(-0.8)} \\
\midrule
ViT-S& 9.6M & 1.29G & 5008 & 63.5 \\
\rowcolor{gray!10}
& 8.1M & 1.25G & 5120 & 61.8 \textcolor{red}{(-1.7)} \\
\rowcolor{gray!10}
\multirow{-2}{*}{\textbf{SKA-ViT-S}}& 9.5M & 1.28G & 5010 & 62.0 \textcolor{red}{(-1.5)} \\
\rowcolor{gray!10}
\textbf{CSKA-ViT-S} & 10.0M & 1.33G & 7220 & \textbf{64.5} \textcolor{darkgreen}{(+1.0)} \\
\bottomrule
\end{tabular}
\end{table}

\paragraph{CIFAR-10/100:}
We evaluated four popular attention-based vision models—CvT, ViT, CaiT, and Swin—by on CIFAR datasets by replacing their attention modules with SKA and CSKA. We then compared their top-1 classification accuracy and computational complexity measured by  the number of images in an inference time of one seconds. Results can be seen in Table \ref{tab:model_comparison_cifar}.
Training and calculation of throughputs are conducted on a single NVIDIA RTX4090 GPU.
We observe that replacing the multi-head attention mechanism in various baseline models with our proposed SKA and CSKA variants yields consistent improvements in terms of both accuracy and computational efficiency. This observation suggests that our proposed methods can perform well in small-scale dataset settings.

\paragraph{ImageNet-1000:}
For ImageNet classification, we used a stronger baseline backbone under the MetaFormer framework. We form our best SKA and CSKA modifications by replacing the stage-2 module (consisting of nine ConvFormer layers) of the CAFormer-S18 with the SKA and CSKA layers. The performance was evaluated in terms of top-1 accuracy in Table \ref{tab:model_comparison_imagenet}. Training and calculation of throughputs are conducted on 8 NVIDIA A100 GPUs. We observe a different behavior compared to the small-scale experiments. Simply replacing all the attention mechanisms in CAFormer with SKA or CSKA does not consistently improve the performance in these larger models, likely due to the higher expressive power required by this dataset. However, a more fine-tuned placement of CSKA improves performance (See Table \ref{tab:ablation3} for the placement ablations). This highlights the effectiveness of the static key attention as a middle-layer architecture in a hierarchical network.

\begin{table}[h]
\scriptsize
\centering
\caption{Comparison of Models on ImageNet. FLOPs are evaluated with a batch size of 128.}
\label{tab:model_comparison_imagenet}
\begin{tabular}{lccl}
\toprule
\textbf{Model} & \textbf{\#Params} & \textbf{FLOPs}  & \textbf{Top-1 acc.} \\
\midrule
\multicolumn{4}{c}{\textbf{ImageNet 224 x 224}} \\
\midrule
RSB-ResNet-50 \cite{res} & 26M & 8.2G  & 79.8 \\
RegNetY-4G \cite{res} & 21M & 8.0G  & 81.3 \\
ConvNeXt-T~\cite{convnext} & 29M & 9.0G  & 82.1 \\
VAN-B2~\cite{van} & 27M & 10.0G  & 82.8 \\
ConvFormer-S18 ~\cite{metaformer} & 27M & 7.8G  & 83.0 \\
DeiT-S ~\cite{deit}& 22M & 9.2G & 79.8 \\
T2T-ViT-14~\cite{t2tvit} & 22M & 9.6G & 81.5 \\
Swin-T~\cite{swin} & 29M & 9.0G  & 81.3 \\
CSWin-T~\cite{cswin} & 23M & 8.6G & 82.7 \\
MViTv2-T~\cite{mvit} & 24M & 9.4G  & 82.3 \\
Dual-ViT-S~\cite{dualvit} & 25M & 9.6G  & 83.4 \\
CoAtNet-0~\cite{coatnet} & 25M & 8.4G  & 81.6 \\
UniFormer-S~\cite{uniformer} & 22M & 7.2G & 82.9 \\
iFormer-S~\cite{iformer} & 20M & 9.6G  & 83.4 \\
CAFormer-S18 (base)~\cite{metaformer} & 26M & 8.2G & 83.3 \\
\midrule
\rowcolor{gray!10}
SKAFormer-S18 & 26M & 8.0G  & 82.3 \textcolor{red}{(-1.0)} \\
\rowcolor{gray!10}
CSKAFormer-S18 & 28M & 8.2G  & 83.4 \textcolor{darkgreen}{(+0.1)}\\
\midrule
RegNetY-16G~\cite{res} & 84M & 31.8G  & 82.2 \\
RegNetY-32G~\cite{res} & 145M & 64.6G & 82.4 \\
ConvNeXt-B~\cite{convnext} & 89M & 30.8G  & 83.8 \\
ConvNeXt-L~\cite{convnext} & 198M & 68.8G  & 84.3 \\
DeiT-B~\cite{deit} & 86M & 35.0G  & 81.8 \\
Swin-B~\cite{swin} & 88M & 30.8G & 83.5 \\
iFormer-L~\cite{iformer} & 87M & 28.0G  & 84.8 \\
MViTv2-L~\cite{mvit} & 218M & 84.2G & 85.3 \\
CoAtNet-3~\cite{coatnet} & 168M & 69.4G  & 84.5 \\
CAFormer-B36 (base)~\cite{metaformer} & 99M & 46.4G & 85.5 \\
\midrule
\rowcolor{gray!10}
SKAFormer-B36 & 97M & 45.6G & 83.9 \textcolor{red}{(-1.6)}\\
\rowcolor{gray!10}
CSKAFormer-B36 & 119M & 48.4G & 85.7 \textcolor{darkgreen}{(+0.2)}\\
\bottomrule
\end{tabular}
\end{table}

\subsection{Comparative results for object detection and instance segmentation}
We conducted experiments on the COCO dataset using both Mask R-CNN and Cascade Mask R-CNN frameworks. The experiments follow a 3× training schedule, a commonly used protocol in object detection benchmarks. Similar to the ImageNet classification model, we replace the stage-2 module with CSKA to evaluate the effectiveness of the new architecture. The computational performance was measured using the number of FLOPs with an input size of $800 \times 1333$ (except MaxViT, which uses $896 \times 896$), and the Frames Per Second (FPS) were recorded on an NVIDIA V100 GPU. Benchmark values of baselines are acquired from \cite{metaformer}. 
The evaluation  includes metrics for object detection (AP\textsubscript{box}, AP\textsubscript{box}\textsuperscript{50}, AP\textsubscript{box}\textsuperscript{75}) and instance segmentation (AP\textsubscript{mask}, AP\textsubscript{mask}\textsuperscript{50}, AP\textsubscript{mask}\textsuperscript{75}). For more details, please refer to the Appendix. As demonstrated in Table \ref{tab:performance_coco}, we observe improved or comparable performance in these tasks compared to the baseline CAFormer with the standard attention mechanism. These results align with the ImageNet classification results and confirm that the CSKA can serve as a good middle-layer architecture in a hierarchical network.

\begin{table}[h]
% \footnotesize
\fontsize{6}{6}\selectfont
\caption{Performance of Object Detection and Instance Segmentation on COCO with Mask R-CNN and Cascade Mask R-CNN}
\label{tab:performance_coco}
\begin{tabular}{lcccccc}
\toprule
\textbf{Backbone} & \textbf{FLOPs} & \textbf{FPS} & \textbf{AP\textsubscript{box} (50 / 75)} & \textbf{AP\textsubscript{mask} (50 / 75)} \\
\midrule
ResNet-50~\cite{deepres} & 376G & 24.0 & 38.2 (60.3 / 41.2) & 34.1 (56.5 / 36.9) \\
ResNet-101~\cite{deepres} & 518G & 17.1 & 40.1 (61.6 / 44.2) & 35.7 (57.6 / 38.5) \\
Swin-T~\cite{swin} & 534G & 19.0 & 46.0 (68.1 / 50.3) & 41.6 (65.1 / 44.9) \\
ConvNeXt-T~\cite{convnext} & 524G & 22.1 & 46.2 (67.9 / 50.8) & 41.7 (65.0 / 44.9) \\
ConvFormer-S18~\cite{metaformer} & 502G & 18.3 & 47.7 (69.6 / 52.3) & 42.6 (66.3 / 45.9) \\
ViT-B~\cite{vit} & 670G & 15.5 & 45.5 (67.4 / 49.8) & 40.9 (63.7 / 44.2) \\
EfficientNet-B7~\cite{efficientnet} & 444G & 21.8 & 45.0 (67.1 / 49.3) & 41.0 (63.9 / 44.1) \\
CAFormer-S18~\cite{metaformer} & 508G & 18.0 & 48.6 (70.5 / 53.4) & 43.7 (67.5 / 47.4) \\
\midrule
CSKAFormer-S18 & 512G & 17.6 & \textbf{48.7} (\textbf{70.7} / 53.4) & \textbf{43.8} (\textbf{67.6} /\textbf{47.5}) \\
\bottomrule
\end{tabular}
\end{table}

\section{Analytic Experiments}
\subsection{Design Ablations}
We explore the impact of various design choices when adapting the traditional self-attention mechanism to use a convolutional static key. We set the baseline as the CAFormer-S18 with its stage 2 substituted with CSKA layers while keeping other components unchanged. We then investigate if the established mechanisms in self-attention remain effective under this new static key configuration. We summarize the results of these design ablations on the ImageNet classification task in Table~\ref{tab:ablation1}.
We observe that the softmax operation plays a crucial role in maintaining performance, even with a static key setup. On the other hand, changes to scaling factors did not significantly influence the final performance.

\begin{table}[h]
\footnotesize
\centering
\caption{Comparison of Models on ImageNet}
\label{tab:ablation1}
\begin{tabular}{lcc}
\toprule
\textbf{Model} & \textbf{Top-1 Accuracy (\%)} \\
\midrule
Baseline & 83.4 \\
\midrule
GeLU & 81.5 \\
ReLU  & 81.1 \\
StarReLU & 81.3 \\
\midrule
No Scaling & 83.3 \\
\bottomrule
\end{tabular}
\end{table}

\subsection{Effect of multi-head}
\begin{figure}[ht]
    \centering
    \includegraphics[width=\linewidth]{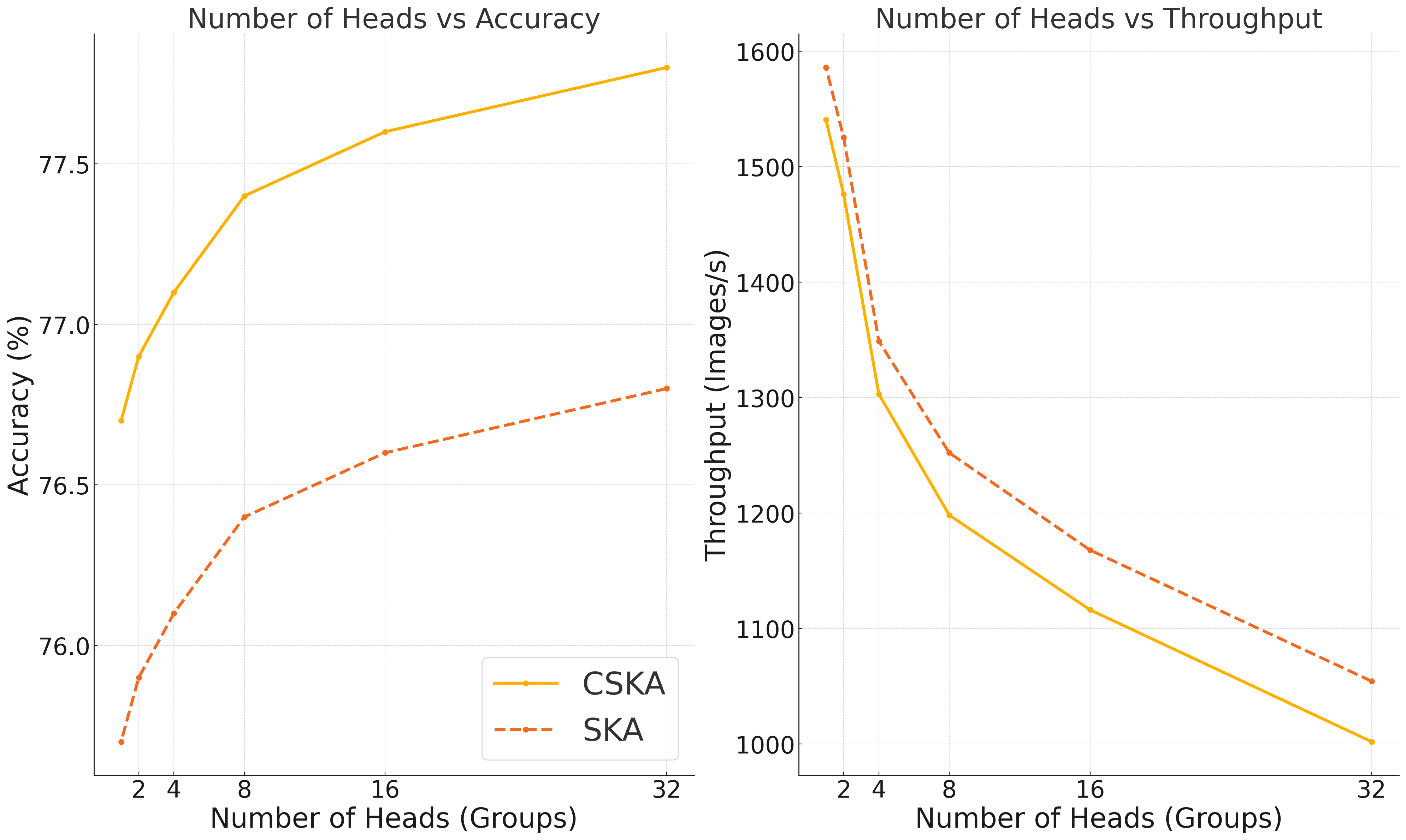}
    \caption{Impact of number of heads (groups) on performance and speed after 160 epochs (full training 310 epochs) training on ImageNet.}
    \label{fig:effect_of_heads}
\end{figure}

Replacing the dynamic key with a static, multi-head network introduces similar computational costs but alters the dynamics of information processing. In this context, multi-headedness is directly applicable to the design of SKA, while for CSKA, it corresponds to the number of groups in the grouped convolution. Similar to standard attention, increasing the number of heads enhances sparsity across embedding dimensions, enabling more specialized and localized interactions within each head or group. To analyze this property, we studied the effect of the number of heads on performance by training the model on ImageNet for 160 epochs. The results are shown in Figure \ref{fig:effect_of_heads}. We observe a behavior consistent with standard attention mechanisms, where a higher number of heads improves the separation and specialization of information. Notably, the trade-off between accuracy and computational efficiency remains evident: as the number of heads increases, the model becomes more expressive, but with a slight increase in computational cost. This consistent pattern underscores the importance of multi-headedness, even in the context of static keys.

\subsection{Effect of position in hierarchical architecture}
Table~\ref{tab:ablation3} shows the ablation results for the placement of SKA and CSKA modules in a four-stage hierarchical architecture based on CAformer-S18, trained on ImageNet-1000 for 160 epochs. The baseline configuration, using depthwise convolution (DW-Conv) in the first two stages and attention mechanisms in the last two, achieves a top-1 accuracy of 77.6\%. Introducing SKA in the second stage yields the same 77.6\%. However, placing SKA in the third stage reduces performance to 77.0\%, and using it in the fourth stage further drops accuracy to 76.7\%. Similarly, CSKA achieves the best performance when placed in the second stage, with a top-1 accuracy of 77.8\%, outperforming both the baseline and SKA. When CSKA is moved to the third stage, accuracy decreases to 77.4\%, and placing it in the fourth stage results in 76.7\%, mirroring SKA’s behavior. These results suggest that SKA and CSKA are particularly effective as middle-stage modules, bridging the gap between feature extraction with DW-Conv and global modeling with attention. The reasons behind their effectiveness in these positions are further analyzed in Sections \ref{sec64} and \ref{sec65}.

\begin{table}[h]
\footnotesize
\centering
\caption{Ablation study of SKA and CSKA module placement in a hierarchical design on ImageNet. DW-Conv stands for depth-wise convolution, and attn stands for standard attention.}
\label{tab:ablation3}
\begin{tabular}{lc}
\toprule
\textbf{Four-stage Configuration} & \textbf{Top-1 acc. (\%)} \\
\midrule
{[}DW-Conv, DW-Conv, attn, attn{]} & 77.6 \\
\midrule
{[}DW-Conv, SKA, attn, attn{]} & \textbf{77.6} \\
{[}DW-Conv, DW-Conv, SKA, attn{]} & 77.0 \\
{[}DW-Conv, DW-Conv, attn, SKA{]} & 76.7 \\
\midrule
{[}DW-Conv, CSKA, attn, attn{]} &\textbf{ 77.8} \\
{[}DW-Conv, DW-Conv, CSKA, attn{]} & 77.4 \\
{[}DW-Conv, DW-Conv, attn, CSKA{]} & 76.7 \\
\bottomrule
\end{tabular}
\end{table}

\subsection{Complexity Comparison}
\label{sec64}

\begin{figure}[h]
    \centering
    \includegraphics[width=\linewidth]{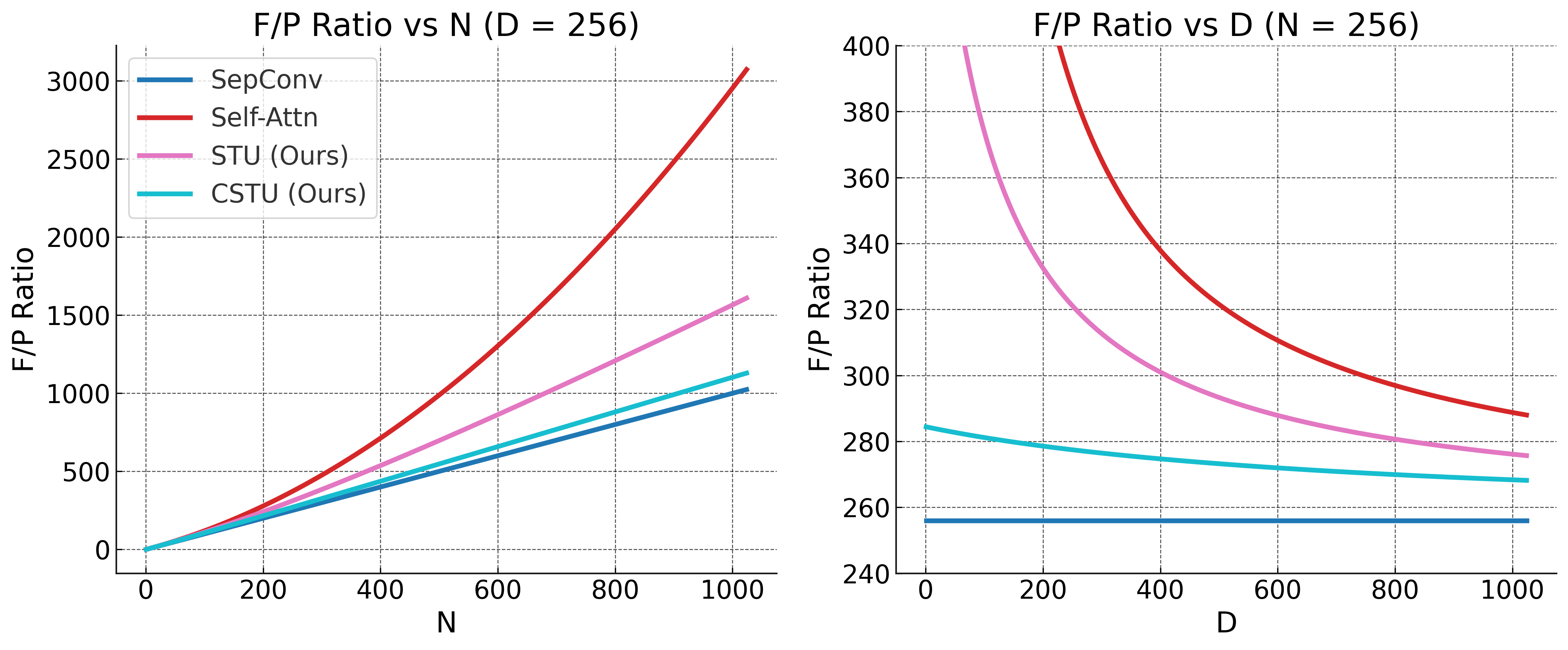}
    \caption{The FLOPs/Parameters ratio vs. spatial dimension (left) and embedding dimension (right). The embedding dimension is set to constant 256 for the left figure, and the spatial dimension is set to 256 for the right figure.}
    \label{fig:fp}
\end{figure}

To study architecture complexity, we provide a comparative analysis of the FLOPs, parameters, and FLOPs-to-parameter ratio for four different token mixing architectures. Each architecture is evaluated on inputs of shape \( N \times D \), where \( N \) is the number of spatial locations (sequence length), and \( D \) is the embedding dimension. The results are summarized in Table~\ref{tab:flops_parameters_ratio} and Figure~\ref{fig:fp}.
\begin{table}[ht]
    \centering
    \scriptsize
    \begin{tabular}{llll}
    \toprule
    \textbf{Model} & \textbf{FLOPs} & \textbf{Params} & \textbf{F/P Ratio} \\
    \midrule
    SepConv & \( N  (9D + 2D^2) \) & \( 9D + 2D^2 \) & \( N \) \\
    Self-Attn & \( N (2N D + 4D^2) \) & \( 4D^2 \) & \( N + \frac{N^2}{2D}\) \\
    SKA (Ours) & \( N  (2N D + 3D^2) \) & \( N D + 3D^2 \) & \( N + \frac{N^2}{N+ 3D}\)\\
    CSKA (Ours) & \( N  (10ND + 3D^2) \) & \( 9ND + 3D^2 \) &  \( N + \frac{N^2}{9N+ 3D}\) \\
    \bottomrule
    \end{tabular}
    \caption{Comparison of FLOPs, Parameters, and FLOPs-to-Parameter Ratio across four models: MLP, Depth-wise Separable Convolution (SepConv), Self-Attention, and our proposed SKA and CSKA.}
    \label{tab:flops_parameters_ratio}
\end{table}

\begin{table}[ht]
    \centering
    \scriptsize
    \begin{tabular}{lcc}
    \toprule
    \textbf{Model} & \textbf{Weight Sharing} & \textbf{Dynamic Weights} \\
    \midrule
    SepConv & Spatially global & No \\
    Self-Attn & No & 2\\
    SKA (Ours) & No & 1    \\
    CSKA (Ours) & Spatially global & 1 \\
    \bottomrule
    \end{tabular}
    \caption{Detailed comparison of Weight Sharing, Dynamic Weights across Depth-wise Separable Convolution (SepConv), Self-Attention, and our proposed models (SKA and CSKA).}
    \label{tab:weight_sharing_dynamic_local_bias_detailed}
\end{table}

As shown in Figure~\ref{fig:fp}, SKA and CSKA have FLOPs/Parameters ratio vs. spatial dimension and FLOPs/Parameters ratio vs. embedding dimension between the curves for self-attention and separable convolution. This intermediate position highlights their balanced trade-off between computational demands and parameter efficiency, making them well-suited for middle layers in hierarchical designs, after depth-wise convolution, and before standard attention mechanisms. The standard attention mechanism is strongly expressive but computationally expensive. In contrast, separable convolution excels at local feature extraction but lacks the global context. SKA and CSKA provide a compromise to benefit from both, efficiently enriching the feature space without the high computational cost of self-attention.
In the hierarchical design, where depthwise convolution captures early local features, SKA and CSKA effectively process more complex spatial relationships before transitioning to the globally-focused self-attention layers. This smooth transition maintains computational efficiency while enhancing representational richness, ensuring an optimal flow of information through the network.

\subsection{Attention Map Comparison}
\label{sec65}

\begin{figure*}
    \centering
    \includegraphics[width=\linewidth]{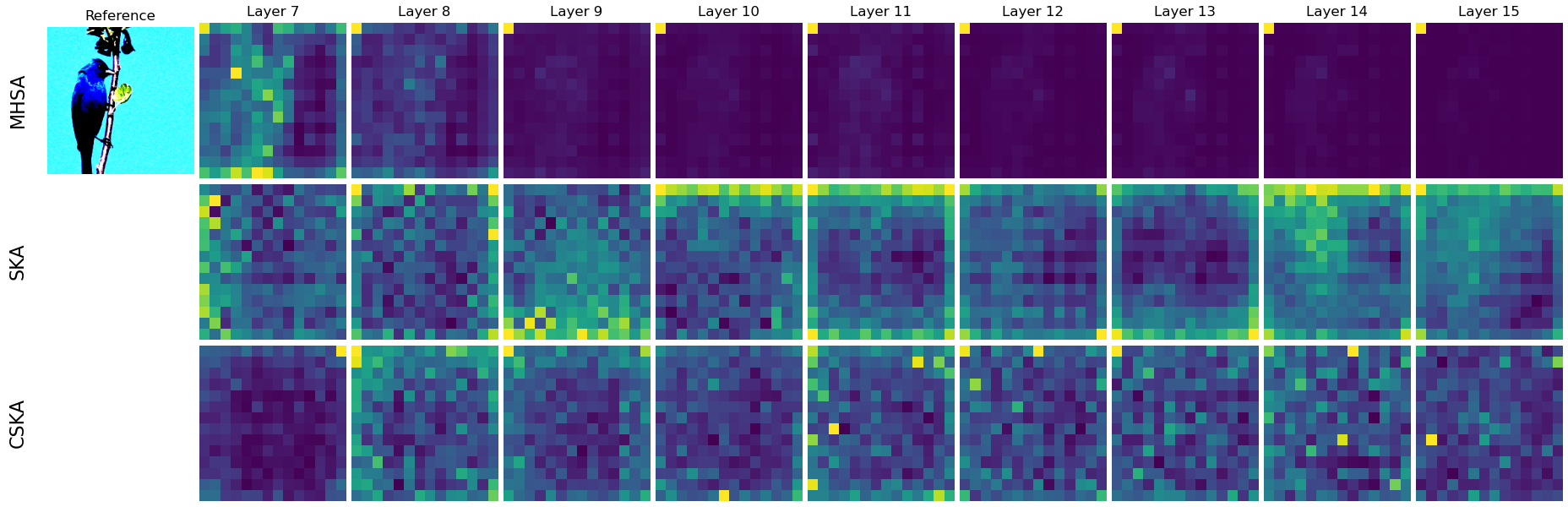}
    \caption{Comparison of SKA and CSKA againt HMSA in terms of interpretability: average attention map over all heads of a reference image at different layers. We choose stage 3 (layers 7 to 15) in the caformer-s18 architecture as the replacement target. Each row corresponds to no replacement (MHSA), replaced with the static key attention (SKA), and the convolutional static key attention (CSKA). SKA and CSKA exhibit vastly different behavior than the MHSA, despite a small modification to the architecture. }
    \label{fig:attnmap}
\end{figure*}
To better understand the behavior of static key attention compared to standard self-attention, we visualize the attention maps generated by both CSKA and SKA mechanisms when used to replace the standard attention in a CAFormer architecture in Figure~\ref{fig:effect_of_heads}. We select a middle stage comprising 9 attention blocks to replace them with the static key attention and convolutional static key attention blocks, then visualize the average attention map over all heads. Our analysis reveals that in the standard CAFormer stage 3, the feature extraction transitions to global information aggregation starting in the third layer (layer 9), in which the attention map indicates the MHSA treats a spatial token as a CLS token and spatial attention is condensed to this single token. But when we replace stage 3 with pure SKA or CSKA, the full stage is dedicated to feature extraction, and the feature maps at the last several layers (layers 14 and 15) start to exhibit similar attention peaks around the object of interest in the reference image. These observations indicate that without a dynamic key, a single dynamic attention-value interaction can learn similar behavior to the standard attention, but takes deeper networks to achieve the same effect. 

However, as we move to the deeper layers, the attention maps in the static key configuration exhibit a more spread-out distribution. This behavior suggests that the static key tends to generalize and capture the broader context in later layers, in contrast to the more precise and localized focus typically observed in standard self-attention. This spread-out behavior in deeper layers might contribute to a better capture of global patterns and spatial relationships, which can be advantageous in complex visual recognition tasks.

% \begin{figure*}[ht]
%     \centering
%     \includegraphics[width=\linewidth, height=200]{attention maps.png}
%     \caption{Attention Map Comparison in Early Layers: Self-Attention vs. Static Key}
%     \label{fig:early_layer_attention}
% \end{figure*}

These visualizations illustrate the evolving nature of attention patterns in the static key mechanism, showing a clear transition from localized focus to more distributed focus as the network depth increases.

\section{Limitations and Future Work}

SKA and CSKA perform well on small-scale datasets such as CIFAR-10/100 but show variability on larger datasets such as ImageNet. Their effectiveness is position-dependent, with the best results when used as intermediate modules, while performance declines when used in early or final stages, limiting architectural flexibility. While SKA is computationally efficient, the grouped convolutions in CSKA can add complexity in deeper models, requiring careful choices of kernel sizes and strides.

Future work could optimize static key mechanisms by fine-tuning the other model configurations beyond the default standard attention. For CSKA, incorporating downsampling functionality directly into the convolutional key could reduce redundancy by eliminating the need for additional dedicated downsampling layers. Observations suggest that static keys effectively capture hierarchical features, warranting further exploration into their impact on downstream tasks and transfer learning. Comparing their learned features with those from standard attention mechanisms could reveal additional representational advantages.

\section{Conclusion}
We introduced Static Key Attention (SKA) and Convolutional Static Key Attention (CSKA) as effective alternatives to the traditional multi-head self-attention (MHSA) mechanism in Vision Transformers. By replacing the dynamic query-key interaction with a trainable static key, these mechanisms streamline the attention process while maintaining or even surpassing the performance of   MHSA under specific conditions. Experiments on image classification, object detection, and segmentation tasks demonstrated that SKA and CSKA achieve competitive results, particularly when used as intermediate modules in hierarchical architectures. Our findings highlight the potential of static key mechanisms and provide insights into how dynamic query-key is not necessary and the softmax activation provides the key expressiveness of the attention mechanism.

{
    \small
    \bibliographystyle{ieeenat_fullname}
    \bibliography{main}
}

\clearpage
\setcounter{page}{1}
\maketitlesupplementary

\section{Extended Experiment Results}
We conducted additional experiments on smaller-scale datasets (\textit{SVHN} and \textit{Fashion-MNIST}) for classification. These experiments extend the findings in the main text to a broader range of datasets. (See Table \ref{tab:model_comparison_svhn_fmnist})

\subsection{Comparative Results for Image Classification}

\begin{table}[h]
\scriptsize
\centering
\caption{Additional results on SVHN and Fashion-MNIST. SKA/CSKA models replace standard attention mechanisms in the baseline models.}
\label{tab:model_comparison_svhn_fmnist}
\begin{tabular}{lcccl}
\toprule
\textbf{Model} & \textbf{\#Params} & \textbf{FLOPs} & \textbf{Tp (image/s)} & \textbf{Top-1 Acc.} \\
\midrule
\multicolumn{5}{c}{\textbf{SVHN}} \\
\midrule
CvT-13 & 19.6M & 0.17G & 6458 & 86.8 \\
\rowcolor{gray!10}
\textbf{SKA-CvT-13} & 19.6M & 0.22G & 6342 & 87.0 \textcolor{darkgreen}{(+0.2)} \\
\rowcolor{gray!10}
\textbf{CSKA-CvT-13} & 18.0M & 0.19G & 6952 & \textbf{87.0} \textcolor{darkgreen}{(+0.2)} \\
\midrule
ViT-S & 9.6M & 1.29G & 5012 & 87.1 \\
\rowcolor{gray!10}
\textbf{SKA-ViT-S} & 9.5M & 1.28G & 5025 & 87.4 \textcolor{darkgreen}{(+0.3)} \\
\rowcolor{gray!10}
\textbf{CSKA-ViT-S} & 10.0M & 1.33G & 7231 & \textbf{88.9} \textcolor{darkgreen}{(+1.8)} \\
\midrule
CaiT-S & 12.7M & 1.42G & 7358 & 86.4 \\
\rowcolor{gray!10}
\textbf{SKA-CaiT-S} & 12.7M & 1.41G & 7100 & 86.8 \textcolor{darkgreen}{(+0.4)} \\
\rowcolor{gray!10}
\textbf{CSKA-CaiT-S} & 12.8M & 1.44G & 7452 & \textbf{89.2} \textcolor{darkgreen}{(+2.8)} \\
\midrule
Swin-T & 26.6M & 1.05G & 5105 & 88.3 \\
\rowcolor{gray!10}
\textbf{SKA-Swin-T} & 26.7M & 0.99G & 5200 & 88.5 \textcolor{darkgreen}{(+0.2)} \\
\rowcolor{gray!10}
\textbf{CSKA-Swin-T} & 26.8M & 1.52G & 4949 & \textbf{88.8} \textcolor{darkgreen}{(+0.5)} \\
\midrule
\multicolumn{5}{c}{\textbf{Fashion-MNIST}} \\
\midrule
CvT-13 & 19.6M & 0.17G & 6478 & 83.0 \\
\rowcolor{gray!10}
\textbf{SKA-CvT-13} & 19.6M & 0.22G & 6332 & 84.4 \textcolor{darkgreen}{(+1.4)} \\
\rowcolor{gray!10}
\textbf{CSKA-CvT-13} & 18.0M & 0.19G & 6962 & \textbf{85.7} \textcolor{darkgreen}{(+2.7)} \\
\midrule
ViT-S & 9.6M & 1.29G & 5000 & 83.2 \\
\rowcolor{gray!10}
\textbf{SKA-ViT-S} & 9.5M & 1.28G & 5013 & 83.6 \textcolor{darkgreen}{(+0.4)} \\
\rowcolor{gray!10}
\textbf{CSKA-ViT-S} & 10.0M & 1.33G & 7242 & \textbf{84.1} \textcolor{darkgreen}{(+0.9)} \\
\midrule
CaiT-S & 12.7M & 1.42G & 7350 & 82.8 \\
\rowcolor{gray!10}
\textbf{SKA-CaiT-S} & 12.7M & 1.41G & 6915 & 83.3 \textcolor{darkgreen}{(+0.5)} \\
\rowcolor{gray!10}
\textbf{CSKA-CaiT-S} & 12.8M & 1.44G & 7407 & \textbf{85.9} \textcolor{darkgreen}{(+3.1)} \\
\midrule
Swin-T & 26.6M & 1.05G & 5105 & 88.1 \\
\rowcolor{gray!10}
\textbf{SKA-Swin-T} & 26.7M & 0.99G & 5200 & 88.0 \textcolor{red}{(-0.1)} \\
\rowcolor{gray!10}
\textbf{CSKA-Swin-T} & 26.8M & 1.52G & 4949 & \textbf{88.2} \textcolor{darkgreen}{(+0.1)} \\
\bottomrule
\end{tabular}
\end{table}

\paragraph{SVHN and Fashion-MNIST:}
SVHN consists of 73,257 training and 26,032 testing images of street-view digits (0-9), with challenging backgrounds. Fashion-MNIST contains 70,000 grayscale images of clothing items across 10 classes, providing a harder alternative to MNIST.

\section{Model and Algorithm}

\paragraph{Small-scale experiment models:}
The four baseline models used in CIFAR-10/100, SVHN, and Fashion-MNIST classification tasks are listed in Table \ref{tab:model_specs_small_consistent}. SKA and CSKA variants replace all attention modules in these models.  

\paragraph{Large-scale experiment models}
We use the CAFormer-S18 and CAFormer-B36 for ImageNet-1K and COCO experiments. The implementations refer to the original paper \cite{metaformer}.

\paragraph{Code implementation:}
\begin{algorithm}[H]
\caption{Static Key Attention (SKA) Mechanism}
\label{alg:ska}
\begin{lstlisting}[language=Python, basicstyle=\ttfamily\scriptsize, keywordstyle=\color{blue}\bfseries, commentstyle=\color{green!50!black}\itshape, breaklines=true]
class SKA(nn.Module):
    def __init__(self, dim, num_tokens, num_heads, qkv_bias=True):
        super().__init__()
        self.num_heads = num_heads
        head_dim = dim // num_heads
        self.q = nn.Linear(dim, dim, bias=qkv_bias)
        self.k = nn.Parameter(torch.randn(num_heads, num_tokens, head_dim))  # Static key
        self.v = nn.Linear(dim, dim, bias=qkv_bias)
        self.proj = nn.Linear(dim, dim)

    def forward(self, x):
        B, N, C = x.shape
        head_dim = C // self.num_heads
        q = self.q(x).reshape(B, N, self.num_heads, head_dim)
        q = q.permute(0, 2, 1, 3)  # [B, H, N, head_dim]
        v = self.v(x).reshape(B, N, self.num_heads, head_dim)
        v = v.permute(0, 2, 1, 3)  # [B, H, N, head_dim]
        k = self.k.unsqueeze(0).expand(B, -1, -1, -1)  # [B, H, num_tokens, head_dim]
        attn = (q @ k.transpose(-2, -1)).softmax(dim=-1)  # [B, H, N, num_tokens]
        x = (attn @ v).transpose(1, 2).reshape(B, N, C)  # Combine heads
        return self.proj(x)
\end{lstlisting}
\end{algorithm}
We provide implementations for both SKA and CSKA mechanisms. These implementations are designed to integrate seamlessly into existing neural network architectures, enabling researchers and practitioners to experiment with their own attention mechanisms.

\begin{itemize}
    \item \textbf{SKA:} Algorithm \ref{alg:ska} implements a static key as a learnable parameter shared across all tokens, simplifying attention computation while maintaining efficiency.
    \item \textbf{CSKA:} Algorithm \ref{alg:cska} combines a static key specifically for the classification token with convolutional keys for spatial tokens, allowing localized information to enhance the attention mechanism.
\end{itemize}

The code is written in PyTorch style. We will provide the full experiment implementation on GitHub.

\begin{algorithm}[H]
\caption{Convolutional Static Key Attention (CSKA) Mechanism}
\label{alg:cska}
\begin{lstlisting}[language=Python, basicstyle=\ttfamily\scriptsize, keywordstyle=\color{blue}\bfseries, commentstyle=\color{green!50!black}\itshape, breaklines=true]
class CSKA(nn.Module):

    def __init__(self, dim, num_heads, num_tokens, dim_head, dropout=0.0):
        super().__init__()
        self.num_heads = num_heads
        head_dim = dim // num_heads

        self.to_qv = nn.Linear(dim, dim * 2, bias=False)

        # Convolutional Key for spatial tokens
        self.to_k = nn.Conv2d(dim, num_heads * num_tokens, kernel_size=3, padding=1, groups=num_heads)

        # Static key for classification token
        self.cls_k = nn.Parameter(torch.randn(num_heads, 1, dim_head))

        self.to_out = nn.Sequential(
            nn.Linear(dim, dim),
            nn.Dropout(dropout)
        )

    def forward(self, x):
      
        B, N, C = x.shape
        head_dim = C // self.num_heads
        qv = self.to_qv(x).chunk(2, dim=-1)  
        q, v = map(
            lambda t: t.reshape(B, N, self.num_heads, head_dim).permute(0, 2, 1, 3), qv
        )  # [B, H, N, head_dim]
        spatial_q = q[:, :, 1:, :]  # Exclude cls token for convolutional keys
        spatial_q_c = spatial_q.permute(0, 3, 1, 2)  # [B, head_dim, H, W]
        k_conv = self.to_k(spatial_q_c).reshape(B, self.num_heads, -1, head_dim)  # [B, H, spatial_tokens, head_dim]

        # Static Key for classification token
        k_cls = self.cls_k.unsqueeze(0).expand(B, -1, -1, -1)  # [B, H, 1, head_dim]

        # Combine convolutional and static keys
        k_combined = torch.cat([k_conv, k_cls], dim=2)  # [B, H, spatial_tokens + 1, head_dim]

        attn = (q @ k_combined.transpose(-2, -1)).softmax(dim=-1)  # [B, H, N, spatial_tokens + 1]

        x = (attn @ v).transpose(1, 2).reshape(B, N, C)  # Combine heads

        return self.to_out(x)

\end{lstlisting}
\end{algorithm}

\begin{table*}[t]
\scriptsize
\centering
\caption{Specifications of small-scale models (ViT-Small, CaiT-Small, Swin-T, CvT)}
\label{tab:model_specs_small_consistent}
\begin{tabular}{lcccc}
\toprule
\textbf{Specification} & \textbf{ViT-Small} & \textbf{CaiT-Small} & \textbf{Swin-T} & \textbf{CvT} \\
\midrule
\textbf{Input Shape}   & $32 \times 32 \times 3$ & $32 \times 32 \times 3$ & $32 \times 32 \times 3$ & $32 \times 32 \times 3$ \\
\midrule
\textbf{Patch Size}    & $4 \times 4$ & $4 \times 4$ & $4 \times 4$ & \begin{tabular}[c]{@{}c@{}}Stage 1: $7 \times 7$ \\ Stage 2: $3 \times 3$ \\ Stage 3: $3 \times 3$\end{tabular} \\
\midrule
\textbf{Embedding Dim} & 512 & 512 & 96 & \begin{tabular}[c]{@{}c@{}}Stage 1: 64 \\ Stage 2: 128 \\ Stage 3: 256\end{tabular} \\
\midrule
\textbf{Depth (Layers)} & 6 & \begin{tabular}[c]{@{}c@{}}Patch: 6 \\ Cls: 2\end{tabular} & \begin{tabular}[c]{@{}c@{}}Stage 1: 2 \\ Stage 2: 2 \\ Stage 3: 2\end{tabular} & \begin{tabular}[c]{@{}c@{}}Stage 1: 1 \\ Stage 2: 2 \\ Stage 3: 10\end{tabular} \\
\midrule
\textbf{Attention Heads} & 8 & 8 & \begin{tabular}[c]{@{}c@{}}Stage 1: 3 \\ Stage 2: 6 \\ Stage 3: 12\end{tabular} & \begin{tabular}[c]{@{}c@{}}Stage 1: 1 \\ Stage 2: 3 \\ Stage 3: 6\end{tabular} \\
\midrule
\textbf{MLP Dimension}   & 512 & 512 & 384 & \begin{tabular}[c]{@{}c@{}}Stage 1: 256 \\ Stage 2: 512 \\ Stage 3: 1024\end{tabular} \\
\midrule
\textbf{Classification Layer Depth} & - & 2 & - & Last stage only \\
\midrule
\textbf{Dropout Rate}    & 0.1 & 0.05 - 0.1 & 0.0 & 0.0 \\
\bottomrule
\end{tabular}
\end{table*}

\end{document}